\documentclass{article}

\usepackage{arxiv}

\usepackage[utf8]{inputenc} 
\usepackage[T1]{fontenc}    
\usepackage{hyperref}       
\usepackage{url}            
\usepackage{booktabs}       
\usepackage{amsfonts}       
\usepackage{nicefrac}       
\usepackage{microtype}      
\usepackage{lipsum}		
\usepackage{graphicx}
\usepackage{natbib}
\usepackage{doi}
\usepackage{amsmath}

\usepackage[final]{changes}

\definechangesauthor[name="Fabian Transchel", color=orange]{FT}
\definechangesauthor[name="Jonathan Schuster", color=blue]{"2"}

\title{maneuverRecognition - a python package for timeseries classification in the domain of vehicle telematics}

\author{\href{https://orcid.org/0000-0000-0000-0000}{\includegraphics[scale=0.06]{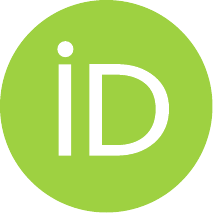}\hspace{1mm}Jonathan Schuster} \\
	Department of Automation and Computer Science\\
	Harz University of Applied Sciences\\
	Wernigerode, Germany \\
	\texttt{schuster.jonathan95@gmail.com} \\
\And
\href{https://orcid.org/0009-0006-7048-353X}{\includegraphics[scale=0.06]{orcid.pdf}\hspace{1mm}Fabian Transchel} \\
	Department of Automation and Computer Science\\
	Harz University of Applied Sciences\\
	Wernigerode, Germany \\
	\texttt{ftranschel@hs-harz.de} \\
}

\renewcommand{\shorttitle}{maneuverRecognition}

\hypersetup{
pdftitle={A template for the arxiv style},
pdfsubject={q-bio.NC, q-bio.QM},
pdfauthor={David S.~Hippocampus, Elias D.~Striatum},
pdfkeywords={First keyword, Second keyword, More},
}

\begin{document}
\maketitle

\begin{abstract}
	In the domain of vehicle telematics the automated recognition of driving maneuvers is used to classify and evaluate driving behaviour. This not only serves \added[id=FT, comment="Präzisierung"]{as a component} to enhance the personalization of insurance policies, but also to increase road safety, reduce accidents and the associated costs as well as to reduce fuel consumption and support environmentally friendly driving. In this context maneuver recognition technically requires a continuous application of time series classification which poses special challenges to the transfer, preprocessing and storage of telematic sensor data, the training of predictive models, and the prediction itself. Although much research has been done in the field of gathering relevant data or regarding the methods to build predictive models for the task of maneuver recognition, there is a practical need for python packages and functions that allow to quickly transform data into the required structure as well as to build and evaluate such models. The maneuverRecognition package was therefore developed to provide the necessary functions for preprocessing, modelling and evaluation and also includes a ready to use LSTM based network structure that can be modified. The implementation of the package is demonstrated using real driving data of three different persons recorded via smartphone sensors.
\end{abstract}


\section{Introduction}

In the context of vehicle telematics, several approaches have already been investigated for automated recognition of driving maneuvers. While previous research has focused strongly on the comparison of different methods for recording the necessary data or the creation of different models to predict or classify driving behaviour \protect\cite{Chan2020,Bejani2020}, there has been a lack in the development of publicly available\added[id=FT, comment="Präzisierung"]{, open source } resources to apply maneuver recognition or even to preprocess data into the required structure for a continuous application of time series classification. In the following, a brief overview on the background of maneuver recognition in the domain of vehicle telematics will be given. Then, the maneuverRecognition Python package is introduced, and the implementation is demonstrated using real-world driving data that has been recorded via smartphone sensors. Afterwards, the results will be observed and discussed. In  conclusion, the usefulness as well as further requirements are going to be evaluated.

\section{Background}

Due to the strong increase in automobile traffic worldwide and the consequent rise of traffic accidents, the relevance of investigating the driving behavior of people has seen increased interest. For this reason, not only traffic and mobility research but also the insurance industry is concerned with the classification of driving maneuvers. Driving behavior can be defined as the way a person responds to both their current state, such as their speed and distance from vehicles ahead, and to changes in their environment, such as traffic or road conditions (\cite{Chan2020}). It is of particular importance which maneuvers are executed, which characteristics they have and how the execution differs depending on the person, road condition or traffic situation. For example, lane changes, evasive maneuvers, speeding, or actions such as turning at intersections can be relevant to differentiate types of drivers (\cite{Wahlstrom2017}).

Maneuver recognition essentially involves the following elements: In most cases, specially developed hardware like a so-called black box (\cite{Wahlstrom2017}) or a smartphone application that records sensors like the accelerometer or the gyroscope is used to capture the required movement data. Subsequently, the collected data is labeled with the respective driving maneuvers and preprocessed to develop predictive models. These models should then be able to classify future data automatically or semi-automatically. Since the recording of driving behavior happens continuously and there are a variety of different driving maneuvers, a multi-class time series classification model that can be used on a continuous stream of data is required. Therefore, an essential difference in the classification of time series data with separated samples (e.g. the classification of music genres based on individual audio files for each song) becomes apparent (\cite{Ehling2020}). Furthermore in this task the class distribution can be extremely unbalanced depending of which maneuvers are investigated and how extensive the given data is. Since risk assessment in insurance cannot rely on observing accidents and near-crashes alone due to their prohibitively scarce incidence, the frequency and incidence of more frequent maneuvers and the relative proportion thereof are of prime concern(\cite{Weidner2017}).

Although there are python packages like pyts by \cite{Faouzi2020} that are specifically designed for the purpose of time series classification, these mainly include functions for feature extraction, the direct application of classification methods on raw time series data or the imputation of missing data.

In order to prepare data so that a model can be trained for maneuver recognition, it is also necessary to perform sliding windowing and divide the data into training and test partitions without introducing data leakage through overlapping windows. When data is collected for the first time and manually labeled with the corresponding maneuver classes for training, these maneuvers are basically available in a form that has isolated windows for every maneuver. However, since later on the models have to predict future driving data, where the length, the beginning and the transitions between the executed maneuvers are unknown, the training process has also to be based on a method where windowing is performed continuously (\cite{Weidner2016}). This leads to overlapping windows and the necessity to use them in the training process in a way that does not create an overlap between the windows of the training data and the testing data. In terms of evaluating multi-class predictive models, there are also few resources available in the existing landscape of time series classification packages. Although confusion matrices are often used for this, if the class distribution is very unbalanced, a more detailed class-wise inspection of the prediction results and performance of the model is necessary.

Especially in the area of maneuver recognition with real world data this is recommended, because transitions between different maneuvers occur seamlessly and can also be interpreted quite differently in real traffic. For example it can easily happen, that performing a left turn on a crossroad will be classified as passing a left curve if the actual traffic situation allows a too similar execution. For these purposes, the maneuverRecognition package contains three different modules for the steps of preprocessing, modeling and evaluation. The classes and functions were first developed during the implementation of the project which is used as a demonstration in section four.

\section{Package content}

The maneuverRecognition package includes three modules, which are related to the different steps in the process of developing a maneuver recognition model.

\subsection{Preprocessing module}

The preprocessing module provides functionalities to transform data into the required structure of time series classification and the usage of recurrent  \added[id=FT, comment="Präzisierung"]{artificial neural }networks. The most important component of this module is the timeseries train test split function, which allows splitting the data into training and testing partitions and then applying a sliding windowing function to each partition separately. Since the use case of maneuver recognition will often require overlapping windows, this function allows to perform a sliding windowing without having any data leakage between the training and testing partitions.

It can also be used to apply robust scaling via the parameter scale. The remove maneuvers function can then be used to balance the class distribution by removing maneuvers that have occurred too rarely or by undersampling maneuvers of majority classes. It is important to not apply this balancing before using the timeseries train test split function since this would lead to a break in the time series data and result in unnatural transitions between the maneuvers that happened before and after the maneuver that is to be removed. As the maneuver types are labeled, the LabelEncoding class can be used to transform the labels into a numeric format. Objects of this class also store the fitted label encoder as well as the given labels, so that an inverse transform function is provided. For using Pytorch the transform to variables function can be applied to transform training and testing data from Numpy arrays to torch Tensor variables.

\subsection{Modelling module}

The modelling module includes an LSTM model structure that can be trained and used for the purpose of time series classification. The basic structure is predefined, but the network is customizable via various parameters. By creating an instance of the class ManeuverModel the model will be initialized. The network structure consists of at least one LSTM layer, which is connected to a linear classifier via two fully connected linear layers. It can be configured to have an arbitrary number of sequential LSTM layers with dropouts in between. The number of hidden features in the connection of the LSTM layers and the first fully connected layer can also be set as well as the number of features in each linear layer.

When training the model, the loss function, optimization algorithm, number of epochs, batch size and device to use can also be configured. The model can be fitted by using the fit model function. To use the model for predictions, the Model object itself includes a predict function. Beside the fit model function the modelling module also provides the separate functions train and test which can generally be used to execute a training or evaluation step. The train, test and fit model functions are not part of the ManeuverModel class in order to enable independent use of the functions with other models and network structures.

\subsection{Evaluation module}

By using the plot training process function of the evaluation module, the validation loss and validation accuracy per epoch can be visualized to examine the training process and assess whether the training has led to over or underfitting. In order to evaluate the model performance the confusion heatmap function can be used with predicted and actual classes of testing data and the corresponding class labels. This function plots a heatmap with confusion matrix values. When the given class distribution is unbalanced these values should be inspected relative to the amount of cases per actual or predicted class individually since the colorization of this regular heatmap takes all fields into account.

With the recall heatmap and the precision heatmap functions the proportion of correctly classified values for each class can be inspected separately in regard to whether a specific class was predicted or a specific actual class is given. The recall heatmap function can be used to inspect the proportion of correctly classified cases per actually given class. This will make the diagonal of the heatmap represent a recall value for each class. It can be read as follows: Given an actual class the row values show the proportion of predictions per class. The precision heatmap function can be used to inspect the proportion of correctly classified cases per predicted class. This will make the diagonal of the heatmap represent a precision value for each class. It can be read as follows: Given cases are predicted to be from a specific class the column values show the proportion of actual class values per class.

\section{Implementation}

In this section, the implementation of the package is demon-strated using real world driving data of a previous research project which has been carried out at the Harz University of Applied Sciences (\cite{Schuster2023}). First, the background of the dataset used is presented. Then the procedural steps of maneuver recognition are performed using the modules of the maneuverRecognition package.

\subsection{Dataset}

\begin{figure}
    \centering
    \includegraphics[width=1\linewidth]{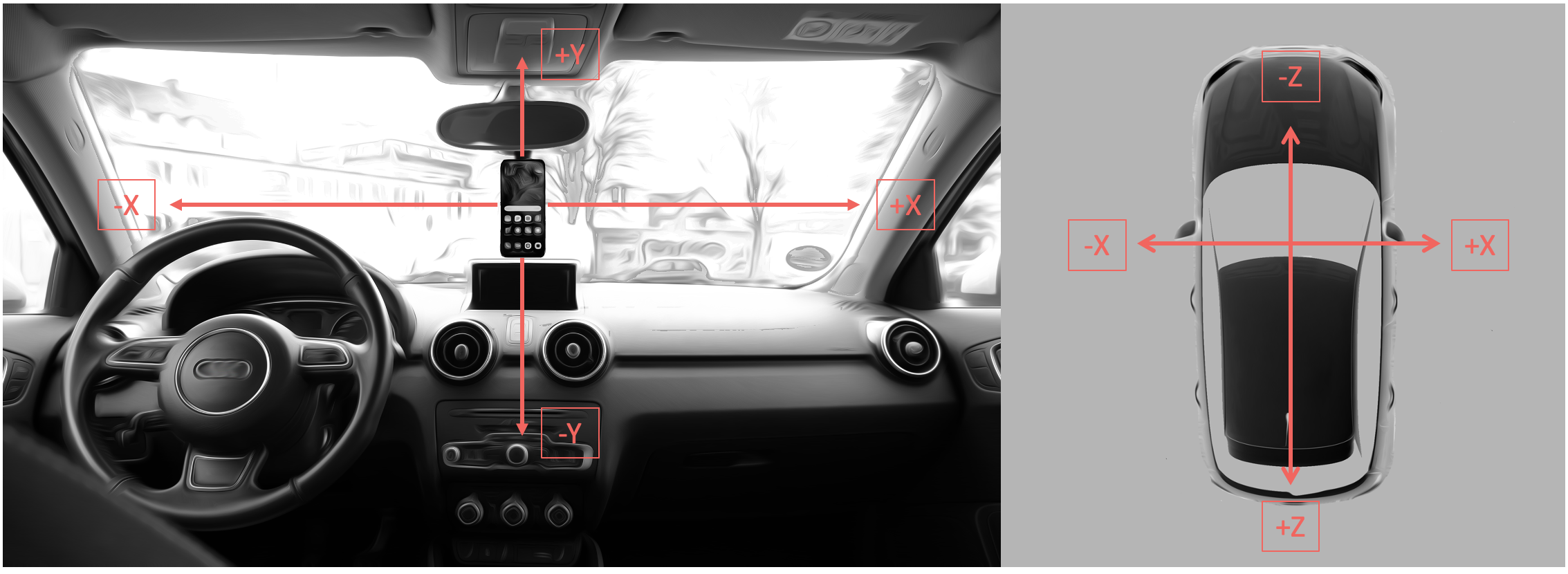}
    \caption{Positioning of the smartphone and marked alignment of the sensor axes.}
    \label{fig:sensor_config}
\end{figure}

The dataset contains sensor data of a smartphone's \textit{accelerometer}, \textit{gyroscope} and \textit{GPS} which were recorded during a total of three trips driven by different persons using different vehicles on the same 60.7\,km route. The route includes multiple road types and passes through urban residential areas, inner city areas, commercial and industrial areas, rural environments and villages. The accelerometer was used to record the acceleration forces along three orthogonal axes in $\frac{m}{s^2}$ without taking the gravitational force into account (\cite{Google2023}). The gyroscope was used to measure \textit{rotational speed} around the same three axes in $\frac{\text{rad}}{s}$. GPS data includes \textit{latitude, longitude, altitude, accuracy} (as an estimated horizontal accuracy radius in meters), and \textit{speed} (in $\frac{m}{s}$). Accelerometer and gyroscope data was updated \replaced[id=FT, comment="Wortverwendung korrigiert"]{every}{at a rate of} two hundred milliseconds, GPS data \replaced[id=FT, comment="Wortverwendung korrigiert"]{every}{at a rate of} three seconds. The most recent sensor data was captured \replaced[id=FT, comment="Wortverwendung korrigiert"]{every}{at a rate of} five hundred milliseconds. The positioning of the smartphone during the recordings and the resulting alignment of the sensor axes can be seen in Figure \ref{fig:sensor_config}. Additionally to the sensor data video footage was captured and used later on to manually label the performed maneuvers as well as the type of road driven.

\subsection{Package application}

The variables that will be used for predicting the previously manually labeled maneuver type are the accelerometer and gyroscope variables for each of the three described axes as well as the GPS speed and the road type. Although the road type was also manually labeled in this data set it can be used since the road type of new driving data can be accessed by matching the GPS coordinates with publicly available road data. In order to meet the requirements of maneuver recognition, the given data will be preprocessed in different ways.

Since maneuver recognition has to be applied on a continuous stream of data and time series classification is often performed by using recurrent neural networks, the data will first be transformed by applying the timeseries train test split function of the \textbf{preprocessing} module. To prevent mixing the data of the three different persons when executing the rolling windowing the function will be applied individually and the transformed data will be combined afterwards.

First of all, the function splits the data into a chosen amount of twenty partitions from which partitions for training and testing data will randomly be sampled. Splitting the data before applying the windowing is necessary in order to prevent data leakage. When selecting the amount of partitions to use for the random sampling the overall size of available data has to be considered since splitting data at some points in time can result in cutting off a maneuver. Using the scale parameter of the function a robust scaling will also be applied. After sampling training and testing data and applying the robust scaling the function executes a sliding window on the training and testing partitions separately. While in the original data each case represents one point in time, this transformation changes the shape of the data so that each case comprises a window of fourteen time steps. With a step size of six time steps between each window and given the capture rate of five hundred milliseconds in the data this leads to a window length of seven seconds and a time between windows of three seconds. For each case the mode of the maneuver type will be used as the target variable value.

With training and testing cases in place the distribution of cases per maneuver type will be inspected. Because of a heavy class imbalance the remove maneuvers function of the \texttt{preprocessing} module will be used to entirely remove some maneuvers that have occurred too rarely (e.g. overtaking) and to undersample maneuvers that are overrepresented (e.g. continuous driving). Removing these maneuvers before applying the time series train test split function would lead to unnatural transitions from the maneuver that happened before to the maneuver after each of the maneuvers which will get removed. The remaining maneuver types that will be used can be seen in Table 1.

The next step in preprocessing the data will be the label encoding of the target variable. Since the class labels will later be used again in visualizing the model performance and generally after using the model for predictions, the \texttt{LabelEncoding} class of the \texttt{preprocessing} module is going to be used to apply the encoding and store both the class labels as well as the encoder object which can be used for an inverse transform later on. Lastly the transform to variables function will be used to transform the training and testing partitions from numpy arrays into pyTorch tensor variables.

The \texttt{modelling} module will then be used to create a pyTorch LSTM model. The base structure of this model is already defined in the package but can still be configured. In this case the model will consist of two LSTM layers with a dropout rate of 0.7 which are connected to two fully connected linear layers with a dropout rate of 0.3. The last layer of the model is going to be a classifier. By using the \texttt{ManeuverModel} class and the texttt{fit} model function of the modelling package the model will be initialized and trained.

In order to evaluate the model at first the progression of the validation loss and validation accuracy during the training process can be inspected by using the plot training process function of the evaluation module. This function shows the values of validation loss and validation accuracy per epoch and can be used to assess whether the amount of training is in proportion to possible over- or underfitting. For evaluating how accurate the model predicts unknown data first the predict function of the model object will be used to predict the classes of the testing data. Then the inverse transform function will be used to transform the predicted values and the true values of the testing partitions target variable to the corresponding class labels.

These could then be used with the confusion \texttt{heatmap} function to evaluate how accurate the model predicts each class by visualizing the amount of cases per actual and predicted class. Since this is an unbalanced multi-class use case, and the color intensity of a regular heatmap takes all fields into account, the distribution will instead be inspected using the precision heatmap and recall heatmap functions (figure 2). In this way the proportion of correctly classified values for each class will be inspected separately in regard to whether a specific class was predicted or a case of a specific class was given. These functions also show a precision and recall value for each class in the diagonal of the corresponding plot.

\subsection{Proof of concept results}

\begin{figure}
    \centering
    \includegraphics[width=1\linewidth]{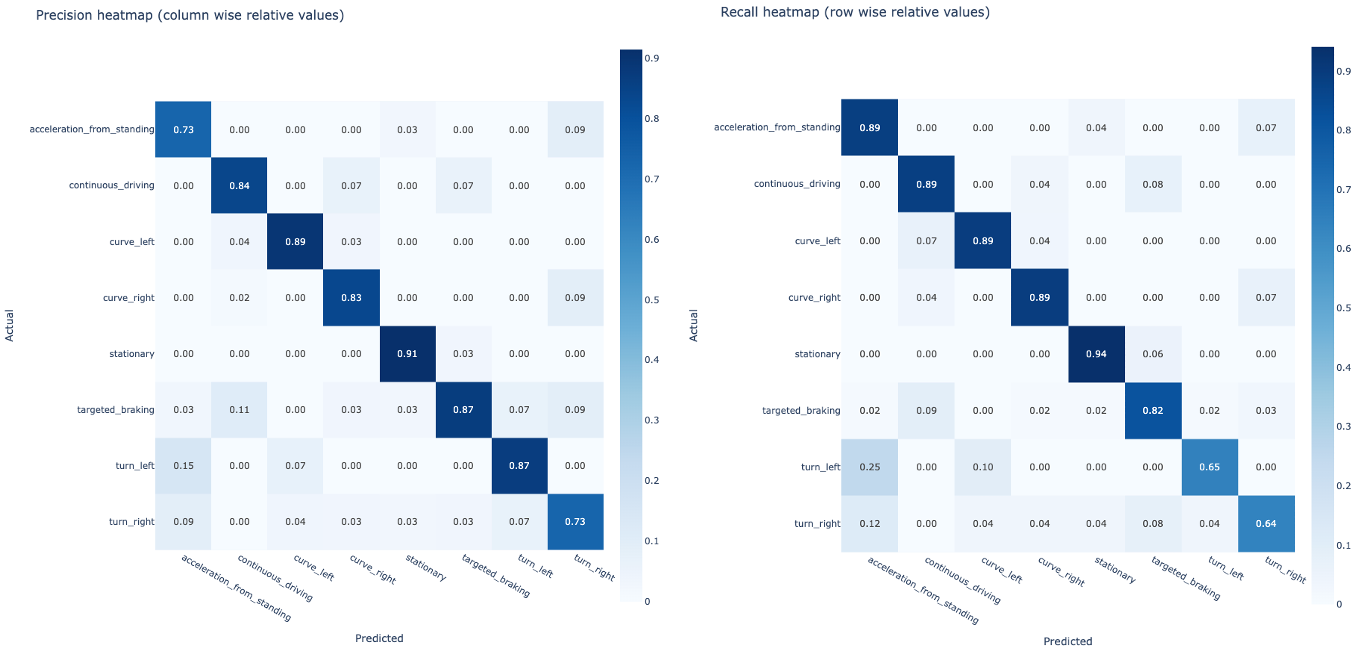}
    \caption{Output of the precision heatmap function (left) and the recall heatmap function (right).}
    \label{fig:precision_recall_contingency}
\end{figure}

Figure \ref{fig:precision_recall_contingency} shows the outputs of the precision heatmap and recall heatmap functions. The lowest precision value is 0.73 for the classes acceleration from standing and turn right while the highest is 0.91 for stationary. Of all predictions as acceleration from standing fifteen percent of cases were actually of class turn left and nine percent actually of class turn right. Of all predictions as continuous driving eleven percent of predicted cases were actually targeted braking maneuvers. The lowest recall value is 0.64 for the class turn right while the highest value is 0.94 for stationary. It can be seen that given the actual maneuver is turn left or turn right a proportion of twenty five and twelve percent of \added[id=FT, comment="Präzisierung"]{respective} cases were incorrectly predicted as acceleration from standing. For actual cases of the class turn left also ten percent of the predictions were incorrectly classified as curve left.

\section{Discussion}

As the demonstration shows, the texttt{maneuverRecognition} package enables quick transformation of recorded time series sensor data and the creation, utilization and evaluation of a predictive model. Nevertheless, limitations in the implementation as well as the evaluation of the presented results have to be considered. First, it should be noted that the used data has been manually labeled and therefore may be fundamentally influenced by inaccuracies both in the accuracy of the labeling and the interpretation of the category system and class definitions. While the results shown in Figure 2 can be considered good for the most part, there are still significant differences between classes. It is important to consider that with the presented texttt{preprocessing} method, the same window length is assigned to all maneuvers and the mode of the maneuver type value within the window is used as the respective target variable. If, as in the given example, there are maneuver types that are executed in very different duration, this can complicate the modeling. Some maneuvers like breaking can often happen in a much shorter time span than a turning maneuver which may includes waiting until oncoming traffic passes. For this reason, both the definition of the maneuvers and their labeling, as well as the selected length of the windows in the preprocessing should be tested and validated. Also the number of time steps between the windows can have a significant impact on the model performance. The model architecture from the modeling module has proven to be feasible in this case and, as mentioned, can be configured via various parameters. Nevertheless, other network structures or model components such as optimization methods or loss functions could also be used with the preprocessed data and lead to different results. 

The evaluation by the proposed precision and recall heatmaps has nevertheless proven to be extremely useful, as it allows a differentiated view of the interrelationships of classes and their predictions as well as the proportion of wrong classifications. When evaluating the model for real-world use, it matters a lot to see if for example a right turn is more likely to be wrongly classified as a right curve or a targeted braking. In real traffic and depending on the class definition different maneuvers could possibly be executed very similarly. This method of evaluation can help to assess whether predictions of a class with a lower precision or recall value are completely off or distributed over classes with possible similarities (e.g. right curve and right turn). Lastly, despite the methods presented, it should be remembered that the quality of collected data, the labeling, and variable selection play the most important role in creating maneuver recognition models. Thus, building a comprehensive open database of driving data in the format proposed would extend this work in ameaningful way toward enabling the scientific community to deepen the understanding of different aspects of driving maneuvers.

The package introduced in this paper provides functions for preprocessing of time series data, the creation of predictive LSTM based classification models and the evaluation of multi class predictions.
Although the package was developed with the intention of maneuver recognition within the domain of vehicle telematics, the functionalities are not restricted to the specific variables or sensor types that have been used in the demonstration. Therefore it can also be utilized in other fields where time series classification is to be performed. In particular when sliding windowing is required, the presented functions can simplify the process of data transformation and reshaping. Nevertheless, there are several aspects to improve and to enhance the usefulness of the package. This includes, for example, a more advanced method for rebalancing the classes in the training and testing datasets.

In the current version of the package, rebalancing is only possible by undersampling or excluding classes although in some cases an oversampling functionality could be needed. In the area of modeling, further model variations or network structures could also be added. LSTM models are only a small fraction of the landscape of recurrent neural networks that can be used for the purpose of maneuver recognition. Since in this domain there are often a large number of classes, further investigation of evaluation methods is also strongly recommended. Although the presented method allows to inspect the quality of predictions in detail on class level, the comparison of different models on this level is difficult. Further development of methods for comparing multi-class predictive models is therefore needed, as well as a more precise assessment of the comparison metrics.

\section*{Acknowledgements}
Fabian Transchel wishes to acknowledge the initial endowment of the Data Science Chair at Harz University of Applied Sciences by E+S Rückversicherung AG, Hanover, Germany.

\bibliographystyle{unsrtnat}
\bibliography{paper}

\end{document}